\documentclass[11pt,a4paper]{article}
\usepackage[nohyperref]{acl2019}
\usepackage{times}
\usepackage{latexsym}
\usepackage[utf8]{inputenc}
\usepackage{booktabs}
\usepackage{tikz}
\usepackage{amsmath}
\usepackage{amsfonts}
\usepackage{enumitem}

\DeclareMathOperator*{\Concat}{Concat}
\DeclareMathOperator{\ReLU}{ReLU}

\usepackage{url}

\aclfinalcopy % Uncomment this line for the final submission
\title{Neural Networks as Explicit Word-Based Rules}

\author{Jindřich Libovický \\
  Charles University, Faculty of Mathematics and Physics \\
  Institute of Formal and Applied Linguistics\\
  Malostranské náměstí 25, 118 00 Prague, Czech Republic \\
  \texttt{libovicky@ufal.mff.cuni.cz} \\
    }
\date{}

\begin{document}
\maketitle
\begin{abstract}
Filters of convolutional networks used in computer vision are often visualized
as image patches that maximize the response of the filter.
We use the same approach to interpret weight matrices in simple architectures for natural language processing tasks.
We interpret a convolutional network for sentiment classification as word-based rules. Using the rule, we recover the performance of the original model.
%
%Further, we analyze simple model for natural language inference and part-of-speech tagging and analyze what words are the networks respond to.
%
\end{abstract}

\section{Introduction}

When using convolutional neural networks (CNNs) for computer vision (CV),
the convolutional filters can be visualized as small image patches that
maximize the response to the filter \citep{krizhevsky2009learning,krizhevsky2012imagenet}. 
Intuitively, the more similar a window of
the image to the filter visualization, the higher the neuron activation is.

In natural language processing (NLP), discrete network inputs are first
embedded into a continuous vector space. The projection that follows
the embedding can be interpreted in a similar way as the filters in CV.
We can retrieve the words whose embeddings have the highest response to the projection.
In this abstract, we use this principle to reconstruct a CNN for sentence
classification using explicit rules.
We present a case study of this approach using models for sentiment analysis.

%\section{Related Word}

\section{CNN for Sentiment Analysis}

The goal of sentiment analysis is to decide if a snippet of text speaks positively or negatively about whatever its content is. We train and evaluate our models on the IMDB dataset 
\citep{mass2011learning} that contains 17k training, 7.5k validation and 25k test 
examples
 with a balanced number of positive and negative examples of movie reviews.

\begin{figure}[t]
    \centering
    \resizebox{\columnwidth}{!}{\includegraphics{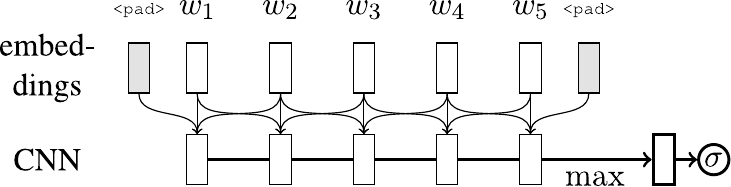}}
    \caption{Architecture used for the sentiment analysis.}
    \label{fig:cnn_arch}
\end{figure}

For our experiments, we use a convolutional network with max-pooling \citep{kim2014convolutional} 
depicted in Figure~\ref{fig:cnn_arch}. We use word embeddings with dimension $d=300$, kernel widths $k$ from 1 up to 5 of $n=500$ filters.

Formally, for a sequence of word embeddings $x_i$ of dimension $d$, the output of the network is:

\begin{equation}\begin{split}
    \mathbf{v} \cdot \Concat_{j=1..k} \left[ \nonumber \max_i \left( \ReLU ( \mathbf{W}_j [x_{i-\frac{j}{2}}, \ldots x_{i+\frac{j}{2}}] ) \right) \right]
\end{split}\end{equation}
where $\mathbf{W}_j \in \mathbb{R}^{d \times n}$ and $\mathbf{v} \in \mathbb{R}^{kn}$ are 
trainable parameters. We apply the sigmoid function over the output and train the network towards the 
cross-entropy loss.

We trained the models until convergence and analyzed the learned weights. Our best
model reached 89\% accuracy, the state-of-the-art result with pre-trained sentence 
representation is 95\% \citep{howard2018universal}.

\section{Model Interpretation}

For each weight vector in each filter, we find words whose embeddings
have the highest dot-product with the weight vector. We 
interpret filters of size 1 as sets of these words.
We interpret kernels of sizes larger
than 1 both either as conjunctions or disjunctions of the neighboring words.
In the conjunction case, we interpret a filter
as a set of n-grams which consists of all combinations of the words extracted from the weight 
vectors. In the disjunction case, we interpret the filters as multiple independent filters
of size 1. Examples of the words extracted from the filters are shown in Table~\ref{tab:filters}.

\begin{table}[t]
\noindent\makebox[\columnwidth]{\rule{\columnwidth}{0.8pt}}
Width 1, weight $8 \cdot 10^{-3}$: \vspace{-7.5pt}
\begin{itemize}[noitemsep]
    \item 1 {\tiny (8.345)}, pointless {\tiny (7.664)}, incohrerent {\tiny (7.270)}
\end{itemize}

\vspace{-4pt}

% Width 1, weight $6 \cdot 10^{-3}$: \vspace{-7.5pt}
% \begin{itemize}[noitemsep]
%     \item   pleasure {\tiny (4.568)}, dissolves {\tiny (4.077)}, delight {\tiny (3.858)}
% \end{itemize}
% 
% \vspace{-4pt}

Width 1, weight $9 \cdot 10^{-3}$: \vspace{-7.5pt}
\begin{itemize}[noitemsep]
 \item perfect {\tiny (12.186)}, brilliant {\tiny (5.268)}, innocent {\tiny (5.040)}
\end{itemize}

\vspace{-4pt}

Width 3, weight $-1 \cdot 10^{-3}$: \vspace{-7.5pt}
\begin{itemize}[noitemsep]
\item yawn {\tiny (7.549)}, incoherent {\tiny (6.338)}, ludicrous {\tiny (6.117)}
\item disappointing {\tiny (4.312)}, acquit {\tiny (4.241)}, appalled {\tiny (4.233)}
\item heather {\tiny (7.362)}, boredom {\tiny (5.949)}, pasolini {\tiny (5.109)}
\end{itemize}

\vspace*{-12pt}

\noindent\makebox[\columnwidth]{\rule{\columnwidth}{0.8pt}}

\caption{Examples of words extracted from the convolutional filters.}
    \label{tab:filters}
\end{table}

We interpret the max-pooling over time as an existential quantifier and thus
the whole sentence representation as asking for the presence of particular words or n-grams, 
i.e., as a set of binary features.

We conduct two experiments with extracted features. First, based on the weight vector
$\mathbf{v}$, we sort the features as contributing to positive
or negative sentiment and label the sentences with the prevailing class.
Second, we train a linear classifier based on the binary features.

The quantitative results of the experiments are shown in Table~\ref{tab:quantitative}.
There is only a minor difference between interpreting the filters as conjunctions 
and disjunctions. This shows that the filters of width 1 are the most important ones
and also that neither of our interpretation of the wider filters is entirely correct.

The experiments with the linear classifier show that when the filters are
interpreted as simple feature extractors, the model performance can be fully
recovered.

\begin{table}[t]
    \centering
    \begin{tabular}{ccccc}
    \toprule
    $k$ &  CNN   & Rules ($\&$) & Rules ($\vee$) & Classifier \\ \midrule
    1   &  83.4  & 83.0         & 83.0           & 86.2 \\
    2   &  85.9  & 82.4         & 81.5           & 86.2 \\
    3   &  87.2  & 81.8         & 81.4           & 86.0 \\
    4   &  87.7  & 81.9         & 81.0           & 86.1\\
    5   &  87.8  & 82.0         & 81.6           & 85.9\\
    \bottomrule
    \end{tabular}
    \caption{Quantiative results of the sentiment classifier and its reconstruction for different kernel sizes $k$.}
    \label{tab:quantitative}
\end{table}

\section{Filter Analysis}

We analyze the part-of-speech of the extracted words.
We computed the most frequent POS tag for each word based on English Web
Treebank \citep{silveira2014gold}.% from Universal Dependencies.
We then computed statistics of the 
most frequent POS tag for words extracted from the network filters.

The statistics are shown in Table~\ref{tab:sentiment_pos_stat}. The most 
frequent POS tag among the extracted word is adjective.
With increasing network capacity, the 
model becomes sensitive to nouns and proper nouns. The proportion
of function words decreases with the increasing kernel size which suggests
that it is unlikely that the filters of large kernel size would capture
more complex phrases.

\begin{table}[t]
    \centering
    \begin{tabular}{lccccc}
\toprule
  &  \multicolumn{5}{c}{Kernel size}  \\ \cmidrule{2-6}
   &  1  &  2  &  3  &  4  &  5 \\ \midrule
 ADJ  &  46.1  &  43.7  &  37.2  &  33.2  &  30.7 \\
 ADV  &  15.8  &  9.2  &  11.2  &  11.9  &  10.7 \\
 NOUN  &  14.6  &  15.8  &  19.1  &  20.5  &  24.6 \\
 VERB  &  11.2  &  16.6  &  19.4  &  20.1  &  19.0 \\
 PROPN  &  1.1  &  6.1  &  5.7  &  8.2  &  8.5 \\
 NUM  &  6.9  &  4.9  &  4.4  &  3.7  &  4.0 \\
 rest  &  4.4  &  3.8  &  2.9  &  2.6  &  2.6     \\
\bottomrule
    \end{tabular}
    \caption{Percentage of POS tags in extracted words for different kernel sizes.}
    \label{tab:sentiment_pos_stat}
\end{table}

We also compared the words extracted from the filters using Opinion Lexicon
\citep{hu2004mining} containing 4.8k words contributing to negative and 2.0k contributing
to positive sentiment. Regardless of the model, approximately 60~\% of the extracted
words appear in the lexicon. If we label the words the sign of the corresponding weight from vector $\mathbf{v}$,
we get precision over 99~\% for both words contributing to the negative and positive 
sentiment with respect to the lexicon.

% \section{Natural Language Inference}
% 
% Natural language inference is theoretically motivate NLP tasks whose
% goal to decide if a pair of sentences is in contradiction, one entail
% the other or are in a neutral relationship.
% We use Stanford NLI dataset \citep{bowman2015large}.
% 
% We used the convolutional architecture with max-pooling for sentence representation 
% as in case of sentiment analysis.
% 
% \todo[inline]{Jak vypada ta architektura, obrazek v tikzu}
% 
% We experimented with similar technique for converting the network
% into a rule-based algorithm, however without any success. We thus
% hypothesize that especially the hidden behaviour of the hidden layers
% is more complex than the sentence representation by CNN and thus cannot
% be easily re-written into rules based on the most prominent weight in
% weight matrix.
% 
% \todo[inline]{Analyzovat POS tagy ve variantach site}
% We analysed the POS tags from the filters in the same as we did with
% 
% \section{Part-of-Speech Tagging} % 
% \todo[inline]{Mozna ho uplne vyhodit}
% \todo[inline]{Co je to za ulohu}
% \todo[inline]{Jak vypada ta architektura, obrazek v tikzu}
% \todo[inline]{Jak vyapada RAN}
% \todo[inline]{Analyzovat POS tagy ve variantach site}

\section{Conclusions \& Future Work}

We showed that the first layer of a CNN
for sentiment analysis can be interpreted
as responding to particular words on input. Using these rules, we fully
reconstruct a model for sentiment classification.

As future work, we would like to extend this approach for more complex 
architectures and other NLP tasks.

\section*{Acknowledgements}

This work has been supported by the grant 18-02196S of the Czech Science Foundation.

\bibliography{references}
\bibliographystyle{acl_natbib}

\end{document}